\documentclass[]{article}
\usepackage{proceed2e}
\usepackage{amssymb,amsmath,graphicx}
\usepackage{subfigure}
\usepackage{mlapa}
\usepackage{algorithm}
\usepackage{algorithmic}

\title{The Latent Bernoulli-Gauss Model for Data Analysis}

\author{Amnon Shashua  \hspace{1cm} Gabi Pragier  \\
\\School of Computer Science and Engineering  \\
Hebrew University of Jerusalem
}

\def\eop  {{\noindent\framebox[0.5em]{\rule[0.25ex]{0em}{0.75ex}}}}
\def\be {\begin{equation}}
\def\ee {\end{equation}}
\def\beas {\begin{eqnarray*}}
\def\eeas {\end{eqnarray*}}
\def\bea {\begin{eqnarray}}
\def\eea {\end{eqnarray}}

\newtheorem{claim}{Claim}

\newcommand{\cond}{\ |\ }

\newcommand{\argmaxx}[1]{\underset{#1}{\mathrm{argmax}} \:}

\newcommand{\argmax}{\mbox{argmax}}

\newcommand{\bfb}{\mbox{\bf b}}
\newcommand{\bfc}{\mbox{\bf c}}

\newcommand{\bfp}{\mbox{\bf p}}
\newcommand{\bfq}{\mbox{\bf q}}

\newcommand{\bfx}{\mbox{\bf x}}

\newcommand{\bfmu}{\boldsymbol{\mu}}

\newcommand{\bflambda}{\boldsymbol{\lambda}}
\newcommand{\bfsigma}{\boldsymbol{\sigma}}
\newcommand{\bfbeta}{\boldsymbol{\beta}}
\newcommand{\bfalpha}{\boldsymbol{\alpha}}

\begin{document}

\maketitle

\begin{abstract}
We present a new latent-variable model employing a Gaussian mixture integrated with a feature selection procedure (the Bernoulli part of the model) which together form a "Latent Bernoulli-Gauss" distribution. The model is applied to MAP estimation, clustering, feature selection and collaborative filtering and fares favorably with the state-of-the-art latent-variable models.
\end{abstract}

\section{Introduction}

We present a new mixture model for  collections of discrete data with applications to clustering through MAP classification, supervised learning, feature selection and collaborative filtering. In the language of text modeling, the algorithm integrates modeling of word frequencies with a feature selection procedure into a single latent class distribution model. The algorithm defines two types of words (i) {\it keywords\/} representing "important" words associated with high frequency appearance, and (ii) all remaining words (not including stop-words which are omitted from consideration). All keywords are "topic specific" modeled by a mixture of Gaussians (one per topic) and all remaining words are considered "topic unspecific" are modeled by a single Gaussian. The decision of which are the keywords of a document is modeled by a latent Bernoulli process --- thus together we have a "Latent Bernoulli-Gauss" (LBG) model.

We present the LBG model in sec.~\ref{sec:bg} and its applications in sec.~\ref{sec:app}. In sec.~\ref{sec:comp} we present  a detailed discussion of the merits of LBG as compared to existing latent-variable models including Mixture-of-Unigrams (MOU) \cite{MoU_Nigam}, probabilistic Latent Semantic Indexing (pLSI) \cite{Hofmann99probabilisticlatent} and Latent Dirichlet Allocation (LDA)~\cite{BleiNJ03}. We conducted a series of experiments on public datasets covering a spectrum of information retrieval applications --- a detailed discussion of experimental results and comparisons to MOU, LDA and pLSI is in sec.~\ref{sec:exp}.

We use the language of text collections throughout the paper, referring to measurements as "word frequencies" and "documents". Nevertheless, the LBG model is general and can be applied (and is applied in sec.~\ref{sec:exp}) to a wide range of data analysis tasks.

\section{The Bernoulli-Gauss Mixture Model}
\label{sec:bg}

Consider a code-book of size $n$ representing the vocabulary of $n$ words in a dictionary. A document is an unordered collection of $N$ words $w_1,...,w_N$ where $w_i\in\{1,...,n\}$. A document $d$ is represented by the $n$ frequencies of word appearances normalized in a proper manner (in text applications we use the term-frequency-inverse-document-frequency (tf-idf) normalization), resulting in $d=(m_1,...,m_n)$ a set of non-negative real numbers.

For a document $d$, we distinguish between a "keyword" which is associated with a high frequency and other low-frequency words of the document. A keyword is another way of saying that the word is "important" for that document. Let $\bfx\in\{0,1\}^n$ be an indicator set where $x_i=1$ if the $i$'th word in the code-book is a keyword and $x_i=0$ otherwise. We assume that the keywords are modeled by a topic-specific Normal distribution whereas all other words are modeled by a topic-unspecific Normal distribution. Let $y\in\{1,...,k\}$ be a random variable representing the $k$ possible "topics" which generated the document $d$. Let $p_{si}$ be the probability  of the $i$'th code-word to be a keyword in the $s=1,...,k$ topic. The Latent Bernoulii-Gauss model $Pr(d\cond y=s,\theta)$ of document $d$ given topic $y=s$ is:

{\small
\be \prod_{i=1}^n \left( p_{si}N(m_i;c_{si},\sigma_{si}^2)\right)^{x_i}
\left((1-p_{si})N(m_i;c_i,\sigma_i^2)\right)^{1-x_i}\label{eq:pdy}
\ee}
where $N(z\ ;\ c,\sigma^2)$ is the Normal distribution $N(c,\sigma^2)$ evaluated at $z$, and $\theta=(\bfp,\bfc,\bfsigma)$ holds the parameters of the model. If the $i$'th code word is a keyword  ($x_i=1$) then the word's frequency $m_i$ is governed by a topic-specific Gaussian distribution $N( c_{si},\sigma_{si}^2)$, otherwise $m_i\sim N(c_i,\sigma_i^2)$ a topic-unspecific Gaussian distribution which we refer to as a "cross Gaussian".
The probability $Pr(d\cond\theta)$ of a document $d$ to be generated by the LBG model is found by the mixture:
\beas
Pr(d\cond\theta) &=& \sum_{s=1}^k Pr(d\cond y=s,\theta)Pr(y=s\cond\theta)\\
&=&\sum_s \lambda_s Pr(d\cond y=s,\theta),
\eeas
where $\sum_s\lambda_s = 1$. Given a training set of documents ${\cal D}=(d_1,...,d_m)$ we wish to fit the model parameters $\theta,\bflambda$ {\it and\/} select the important code words for each document, i.e., estimate ${\cal X}=(\bfx_1,...,\bfx_m)$ where $\bfx_j\in\{0,1\}^n$ is the keyword indicator set associated with $d_j$. We alternate between two procedures: (i) Maximum-Likelihood (ML) estimation of $\{\theta,\bflambda\}$ given $\cal X$ and, (ii) a procedure for estimating $\cal X$ given $\{\theta,\bflambda\}$.

The ML estimation of $\{\theta,\bflambda\}$ given an i.i.d. training set $\{{\cal D},{\cal X}\}$ takes the form:
\beas
&&\max_{\theta} \sum_{j=1}^m \log Pr(d_j\cond \theta,\bfx_j)\\
&& =\max_{\theta,\bflambda} \sum_{j=1}^m\log\left(\sum_{s=1}^k \lambda_s Pr(d_j\cond y_j=s,\theta,\bfx_j)\right)
\eeas
where $Pr(d_j\cond y_j=s,\theta,\bfx_j)$ is given by:
$$\prod_{i=1}^n \left( p_{si}N(m_{ji};c_{si},\sigma_{si}^2)\right)^{x_{ji}}
\left((1-p_{si})N(m_{ji};c_i,\sigma_i^2)\right)^{1-x_{ji}}.$$
Using the Expectation-Maximization (EM) iterative update \cite{Dempster-EM}, the following auxiliary function is optimized during the M-step:
$$\max_{\theta,\bflambda}\sum_{j=1}^m\sum_{s=1}^k \mu^{(t)}_{sj}\log\left(\lambda_sPr(d_j\cond y_j=s,\theta,\bfx_j)\right),$$
where $\mu^{(t)}_{sj}=Pr(y_j=s\cond d_j,\bfx_j,\theta^{(t)})$ is the posterior probability given the parameters at iteration $(t)$. Optimizing over the auxiliary function at step $(t)$ introduces an update rule for $\theta,\bflambda$:
\bea
\lambda_s &\leftarrow& \frac{1}{m}\sum_{j=1}^m\mu^{(t)}_{sj},\ \
p_{si} \leftarrow \frac{1}{\sum_j \mu^{(t)}_{sj}} \sum_{j=1}^m \mu^{(t)}_{sj} x_{ji} \label{eq:up-l}\\
c_{si} &\leftarrow& \frac{1}{\sum_j \mu_{sj}^{(t)}x_{ji}}\sum_{j=1}^m \mu_{sj}^{(t)}x_{ji}m_{ji}\\
\sigma^2_{si} &\leftarrow& \frac{1}{\sum_j \mu_{sj}^{(t)}x_{ji}}\sum_{j=1}^m \mu_{sj}^{(t)}x_{ji}(m_{ji}-c_{si})^2\label{eq:up-sig}
\eea
The parameters $c_i,\sigma_i^2$ of the cross-Gaussians are estimated directly from $\cal D,X$ since they do not depend on the choice of topics. The posteriors are updated during the E-step via application of the Bayes rule:
\be
\mu^{(t+1)}_{sj}\propto \lambda_s^{(t)} Pr(d_j\cond y_j=s,\bfx_j,\theta^{(t)}),\label{eq:mu}
\ee
where $\propto$ stands for equality up to normalization, i.e., $\sum_s \mu_{sj}=1$.

The estimation of $\cal X$ given the data $\cal D$ and the current estimation of parameters $\{\theta,\bflambda\}$ is based on the following analysis. Consider Natural numbers $q_s\in{\cal N}$, $s=1,...,k$, representing the number of important code words associated with topic $s$. The expected number of important code words $g_j$ in document $d_j$ is given below:
\be
g_j = \sum_{i=1}^nx_{ji} = \sum_{s=1}^k \mu_{sj}q_s.\label{eq:g}
\ee
In other words,  the indicator set $\cal X$ is fully determined by $q_1,...,q_k$ and the posteriors $\mu_{sj}$ (which are estimated during the EM step above). The indicator $\bfx_j$ for document $d_j$, for instance, is defined by the top $g_j$ highest frequency code words. Our task, therefore,  is to derive a procedure for estimating $q_1,...,q_k$ given the parameters $\theta,\bflambda$ and $\bfmu$ estimated during the EM steps.

We will begin by establishing an algebraic constraint between $\bfq=(q_1,...,q_k)$ and the parameters $\bfp,\bfmu$:
\begin{claim}
\label{claim1}
Let $\bfb=(b_1,...,b_k)$ defined by $b_s = (\sum_j \mu_{sj})(\sum_ip_{si})$ for $s=1,...,k$, and let $U$ be an $k\times m$ matrix holding the posteriors, $U_{sj}=\mu_{sj}$. Then,
\be
\bfb = UU^\top\bfq\label{eq:uuq}
\ee
\end{claim}
{\bf Proof:\ } consider the formula representing the expected number of important words for a document of topic $s$:
$$E_s = \frac{1}{\sum_j \mu_{sj}}\sum_{j=1}^m\mu_{sj}g_j.$$
On the other hand, clearly, $E_s = \sum_i p_{si}$ since $p_{si}$ is the probability that the $i$'th code word is important for documents of topic $s$. Substituting the definition of $g_j$ from eqn.~\ref{eq:g}, we obtain:
$$(\sum_j \mu_{sj})(\sum_ip_{si}) = \sum_{j=1}^m \mu_{sj}\sum_{r=1}^k \mu_{rj}q_r,$$
where the right hand side is the $s$'th coordinate of $UU^\top\bfq$. \eop

The conditional-independence assumption $w_i\bot w_j\cond y$  (Naive-Bayes) creates "over-confident" posteriors, i.e., $\mu_{sj}\rightarrow \{0,1\}$ --- a well-known by-product (or side-effect) of the Naive Bayes assumption (see \cite{Domingos97} for a discussion). As a result, the constraint $UU^\top\bfq=\bfb$ is simplified considerably: $UU^\top \approx diag(\delta_1,...,\delta_k)$, where $\delta_s\approx \sum_j\mu^2_{sj}\approx \sum_j\mu_{sj}$. Eqn.~\ref{eq:uuq}, therefore, reduces to:
\be
\sum_{i=1}^np_{si}=q_s,\label{eq:bq}
\ee
for $s=1,...,k$. Eqn.~\ref{eq:bq} is not an effective update rule for setting $q_1,...,q_k$ because (i) there is no built-in drive to generate a sparse $\bfp$, which as a result, a large number of small-valued entries in $\bfp$ will inflate the value of $q_s$, and (ii) once entries of $\bfmu$ settle on $\{0,1\}$ values, the indicator set $\cal X$ will remain fixed.

A more effective use of Eqn.~\ref{eq:bq} is to to set $q_s$ as the top number of entries in $\bfp$:
$$q_s^{(t+1)} = |\{i\ :\ p_{si}\ge T_s^{(t)}\}|,$$
for some, iteration dependent, threshold $T_s$. In the following section we use a similar analysis to derive the value of $T_s$ which will conclude the Bernoulli-Gauss mixture algorithm.

\subsection{Update Rule for $q_1,...,q_k$}

\begin{algorithm*}
\caption{Bernoulli-Gauss Mixture}
\label{alg}
\begin{algorithmic}
\STATE {\bfseries Input:} Given a training set of documents ${\cal D}=(d_1,...,d_m)$ we wish to fit the model parameters $\bflambda$ and $\theta=(\bfp,\bfc,\bfsigma)$ for $k$ topics and the Natural numbers $q_1,...,q_k$ of top ranking (by tf-idf) words per topic.
\STATE {\bfseries Initialization:} Set initial values $\bflambda^{(0)},\theta^{(0)},\bfq^{(0)}$. Set the indicators ${\cal X}^{(0)}$ from ${\cal D}$ and $\bfq^{(0)}$, i.e., $x_{ji}=1$ if the if-idf value $m_{ji}$ is among the top $(1/k)\sum_s q_s^{(0)}$ entries in $d_j$. Set $t=0$.
\REPEAT
\STATE $t\leftarrow t+1$
\STATE  Update the posteriors $\mu_{sj}^{(t)}$ according to Eqn.~\ref{eq:mu} for $j=1,...,m$ and $s=1,...,k$.
\STATE  Update $\bflambda^{(t)},\bfp^{(t)},\bfc^{(t)},\bfsigma^{(t)}$ using Eqns.~\ref{eq:up-l}-\ref{eq:up-sig} and then update the cross-Gaussians.
\STATE  Set $\bfq^{(t)}$ using eqn.~\ref{eq:up-q}.
\STATE  Set ${\cal X}^{(t)}$: $x_{ji}=1$ if the if-idf value $m_{ji}$ is among the top $\sum_s\mu_{sj}^{(t)} q_s^{(t)}$ entries in $d_j$, for $i=1,...,n$ and $j=1,...,m$.
\UNTIL{$\sum_{s=1}^k \left( q_s^{(t)}-\sum_ip^{(t)}_{si}\right)^2 < \epsilon$}
 \end{algorithmic}
\end{algorithm*}
Let $q_s^*$ be the (unknown) ground truth value for $q_s$. Since $g_j$ (eqn.~\ref{eq:g}) is the number of keywords in document $d_j$, the probability that {\it a keyword\/}  will be selected in
$d_j$, conditioned by topic $s$, is $\min\{g_j/q_s^*,1\}$. The probability that a keyword will be selected in $d_j$ {\it and\/} the topic is $s$ is a random variable with a Bernoulli distribution with the probability of "success":
$\mu_{sj}\min\{g_j/q_s^*,1\}$.
The expected number of times a keyword is selected over the corpus of $m$ documents of topic $s$ is the sum of expectations of $m$ Bernoulli trials:
$$\sum_{j=1}^m \mu_{sj}\min\{g_j/q_s^*,1\}.$$
On the other hand, the expected number of times the $i$'th  code-word (not necessarily a keyword) is selected in documents of topic $s$ is: $m\lambda_sp_{si}$.
As a result, for the $i$'th code-word to be a keyword for a document of topic $s$ the following condition must be satisfied:
$$
m\lambda_sp_{si} \ge \sum_{j=1}^m \mu_{sj}\min\{g_j/q_s^*,1\}
\ge \frac{1}{n}\sum_{j=1}^m \mu_{sj}g_j,
$$
where the first inequality is due to the rhs being a lower bound for a word to be a keyword, and the latter inequality is due to $1\le q_s^*\le n$. After rearranging terms and substituting eqn.~\ref{eq:g} for $g_j$ we obtain:
$$q_s^{(t+1)} = \left |\left \{i\ :\ p_{si}\ge \frac{1}{n\sum_j\mu_{sj}}\sum_{j=1}^m \mu_{sj}\sum_{r=1}^k \mu_{rj}q_r^{(t)}\right\}\right |.$$
Note that the right-hand side is the $s$'th coordinate of $UU^\top\bfq$ scaled by $1/(n\sum_j\mu_{sj})$. Given that the posteriors $\mu_{sj}$ approach $\{0,1\}$ values, the condition above reduces to:
\be
q_s^{(t+1)} = \left |\{i\ :\ p^{(t+1)}_{si}\ge\frac{1}{n}q_s^{(t)}\}\right |.\label{eq:up-q}
\ee
To conclude, the Bernoulli-Gauss mixture algorithm is summarized in Alg.~\ref{alg}. The stopping criteria is when Claim~\ref{claim1} is satisfied, but in practice it is sufficient to satisfy its reduced form  eqn.~\ref{eq:up-q}.


\subsection{Evaluating the Model on Novel Documents}
\label{sec:eval}

Given a new document $d=(m_1,...,m_n)$, where $m_i$ is the frequency (tf-idf) of the $i$'th code-word in the document,  we wish to evaluate the probability  $Pr(d)$ of $d$ to arise from the model, and the posteriors $\mu_s(d) = Pr(y=s\cond d)$ which provide classification (topic assignment) information. A necessary ingredient in those calculations is the estimation of the keyword indicator set $\bfx\in\{0,1\}^n$ for document $d$.
To estimate $\bfx$ associated with the novel document $d$ we perform the following steps:

\noindent{\bf 1.\ }For $s=1,...,k$: (i) define $\bfx_s$ as the indicator set defined by the top $q_s$ code words in $d$, (ii) compute $\hat\mu_s\propto\lambda_sPr(d\cond y=s,\bfx_s)$ where $Pr(d\cond y=s,\bfx_s)$ is defined in eqn.~\ref{eq:pdy}.\\[1mm]
\noindent{\bf 2.\ }Set $\bfx$ as the top $(1/\sum_s\hat\mu_s)\sum_s\hat\mu_sq_s$ code words in $d$.

Once $\bfx$ is estimated, one can readily compute the posterior
$\mu_{s}\propto \lambda_s Pr(d\cond y=s,\bfx,\theta)$, $s=1,...,k$.
and $Pr(d)$ from:
$Pr(d)=\sum_{s=1}^k \lambda_s Pr(d\cond y=s,\bfx,\theta)$.

\subsection{Applications of the Model}
\label{sec:app}

The Bernoulli-Gauss mixture model can be used in a number of ways and for different data analysis applications, as described below:

\noindent{\bf Clustering:\ }
given documents $d_1,...,d_m$, cluster them into $k$ classes. Moreover, given a novel document $d$ determine its class association. The posteriors $\mu_{sj}$ for $d_j$ and class $s$ provide the class assignment of document $d_j$. Since posteriors are "over-confident" due to the Naive Bayes assumption, the assignment is "hard" in practice. For a new document $d$, the posteriors $\mu_s$ (see Sec.~\ref{sec:eval}) provide the class assignments for $s=1,...,k$.

\noindent{\bf Supervised Inference:\ } given a training set of documents with class labels in the set $\{1,...,h\}$ we wish to determine the class membership of a given novel document. A possible approach is to estimate a LBG model separately for each class producing the model parameters  $\bflambda_l,\theta_l,\bfq_l$, $l=1,...,h$, and then choose the class with the highest probability: $\argmax_l Pr(d\cond \bflambda_l,\theta_l,\bfq_l)$.

\noindent{\bf Feature Selection:\ }
we can use the Bernoulli-Gauss mixture model for selecting features. The selection criteria is based on $p_{si}$ which is the probability that the $i$'th code word (feature) is a keyword  for topic $s$. We "de-select" a feature $i$ if $p_{si}<\delta$ for some threshold $\delta$ for all $s=1,...,k$, i.e., a feature that is {\sl not\/} a keyword in {\sl all\/} topics is removed from the set of selected features. In Sec.~\ref{sec:exp} we apply the feature selection scheme above as a filter for Support-Vector-Machine (SVM) classification and for K-means clustering.

\noindent{\bf Collaborative Filtering:\ }
there are applications where the indicator set $\bfx\in\{0,1\}^n$ is known, and moreover when $x_i=0$ the frequency of the $i$'th code word $m_i$ is unknown. Collaborative Filtering (CF) is an example of this class of applications where $d=(m_1,...,m_n)$ is a list of discrete movie ratings with $m_i\in\{1,...,5\}$ (stars), of an individual. Each individual rates some of the movies, thus $x_i=1$ for movies being rated and $x_i=0$ otherwise.  Given a subset of ratings made by a new individual, the task of CF is to predict movie ratings which were not part of the original subset.

In this case, the cross-Gaussians are dropped from the model, i.e.,
\be
Pr(d\cond y=s)=\prod_{i=1}^n \left( p_{si}N(m_i;c_{si},\sigma_{si}^2)\right)^{x_i}
(1-p_{si})^{1-x_i}.\label{eq:cf}
\ee
From the training ratings $\{d_j,\bfx_j\}$, $j=1,...,m$, we estimate the model parameters $\theta,\bflambda$ using Eqns.~\ref{eq:up-l}-\ref{eq:up-sig} (there is no need to estimate $q_1,...,q_k$ since the indicator sets are known). We are given a new rating $\{d,\bfx\}$ where $d=(m_1,...,m_n)$ and $x_i=1$ when $m_i>0$. Let $i\in\{1,...,n\}$ be a movie we wish to predict its rating by the individual $d$. Similarly to the "Forced Prediction" protocol~\cite{Breese98empiricalanalysis}, we wish to estimate the probability $Pr(m_{i}=t\cond d,\theta)$ for $t=1,...,5$. We start by setting $x_{i}=1$ (originally it was zero):
$$Pr(m_{i}=t\cond d)=\sum_{s=1}^k Pr(m_{i}=t\cond y=s)Pr(y=s\cond d),$$
where
$$Pr(m_{i}=t\cond y=s)=N(t;c_{si},\sigma_{si}^2),$$
and the posterior $Pr(y=s\cond d)\propto\lambda_sPr(d\cond y=s)$ is estimated through eqn.~\ref{eq:cf}. The movie rating prediction $t^*$ is found by: $t^*=\argmaxx{t}Pr(m_{i}=t\cond d)$.

\section{Relationship with Other Latent Variable Models}
\label{sec:comp}

On a simplistic level,
the Bernoulli-Gauss mixture model can be viewed as a Gaussian mixture model integrated with a feature selection procedure (the Bernoulli part of the model). On a deeper level, however, there are subtleties that have to do with the positioning of LBG with respect to MOU, LDA and pLSI and specifically the manner in which LBG is a {\it generative\/} model like MOU and LDA, which we will describe below.

One  difference is that LBG models the frequency of a code word (per topic) as a Gaussian whereas MOU, LDA and pLSI model the probability of appearance of code-words as a multinomial --- which at the limit  are really the same, as described next. Let $\beta_{si}=Pr(w=i\cond y=s)$ be the probability of drawing the $i$'th code-word given the $s$'th topic. The number of appearances $m_i$ of the $i$'th code-word in a document is governed by a Binomial distribution $m_i \sim Bin(N,\beta_{si})$ where $N$ is the number of words in the document. By the De-Moivre-Laplace theorem, as $N\rightarrow\infty$, $m_i\sim {\cal N}(N\beta_{si},N\beta_{si}(1-\beta_{si}))$. Therefore, in practice since the number of words $N$ is a document is typically large,  the estimated means $c_{si}$ in the LBG model are equal to $N\beta_{si}$ in the multinomial models.

The De-Moivre-Laplace argument above is also relevant for the justification of a Gaussian distribution as a model of word frequencies (or any other non-negative data). It implies that the probability of a negative value (in the generative sense) is vanishingly small. Successful attempts in using Gaussian mixtures in non-negative numerical contexts, such as for collaborative filtering, include \cite{CF_Hoff03}. In practice we have not observed any problematic issue with a Gaussian modeling and our experimental reports across a number of application domains (text analysis included) make that point as well.

It will be convenient, in this section, to represent a document $d=(w_1,...,w_N)$ by the (unordered) set of words $w_i\in\{1,...,n\}$ taking values from a vocabulary of $n$ code-words. We will begin the discussion with the comparison between the MOU model   and LBG. A document is generated by the  MOU model by a draw from a mixture of multinomials as follows. A topic is drawn by tossing a $k$-faced die whose faces have probabilities $\lambda_s=Pr(y=s)$.  A word is drawn by the toss of an n-faced die where we have $k$ such dice each representing a topic $s=1,...,k$, with $\beta_{si}$ (as defined above) representing the probability of the $i$'th face of the $n$-face word-die associated with topic $s$.
The $N$ words of a document are generated by (i) draw a topic $s$ by tossing the $k$-faced topic-die, then repeat $N$ times: (ii) draw a code-word by tossing the $s$'th word-die. In formal language,
$$Pr(d)=\sum_{s=1}^k \lambda_s\prod_{i=1}^n\beta_{si}^{m_i}.$$
The model parameters $\bflambda,\bfbeta$ can be estimated by the EM algorithm. The MOU model is simple and very popular in text analysis circles. However, it has a number of drawbacks which have served as a catalyst for introducing new algorithms, notably pLSI and LDA. The notion that all code-words appearance is governed by  the choice of a single topic is too simplistic. First, there are code-words which have a low probability of appearance in {\sl all\/} topics, i.e., are essentially topic-independent, yet are not stop-words. These words undergo "starvation" in the MOU model as they almost never have a chance to be appear in a document generated by MOU. Second, polysemy --- the coexistence of multiple meanings for a code-word --- is not modeled by MOU. Consider a document $d$ and a code-word $w$. In MOU the posterior probability $Pr(y=s\cond w,d)$ is independent of $d$:
$$Pr(y=s\cond w,d)\propto Pr(w\cond y=s)Pr(y=s),$$
therefore it is not possible to convey multiple meanings for the code-word $w$ as a function of other words in the document $d$.

In the LBG model, the single topic assumption applies only to a selected set of code-words, whereas all other code-words are governed by a topic-unspecific distribution. The manner in which this principle plays in a generative model is described formally as follows:
$$Pr(d,\bfx)=\sum_{s=1}^k \lambda_s\prod_{r=1}^NPr(w_r,x_r\cond y=s),$$
where
$$Pr(w_r=i,x_r\cond y=s)\propto\left\{\begin{array}{ll}
p_{si}\frac{1}{N}c_{si} & if\ x_i=1\\
(1-p_{si})\frac{1}{N}c_{i}& if\ x_i=0
\end{array}\right\}
$$
In other words, the $N$ words of a document $d$ are generated through the following steps:
\begin{itemize}
\item Draw a topic $s$ by tossing the $k$-faced topic-die.
\item Toss $n$ coins with biases $p_{si}$, $i=1,...,n$ to draw the indicator vector $\bfx\in\{0,1\}^n$.
\item Create a $n$-faced word-die by setting $\hat\beta_{si}$  to $(1/N)c_{si}$ if $x_i=1$ or to $(1/N)c_{i}$ if $x_i=0$. The probability $\beta_{si}$ of the $i$'th face of the word-die is $(1/Z)\hat\beta_{si}$ where $Z$ is a normalization factor such that $\sum_i\beta_{si}=1$.
\item Repeat $N$ times: draw a word from the word-die constructed above.
\end{itemize}

In other words, in the LBG model the word-die is generated per document not only on the basis of the topic selection  but also {\it based on the selection of keywords}. With regard to polysemy,  the posterior probability $Pr(y=s\cond w,d)$ now depends on $d$:
$$Pr(y=s\cond w,d)\propto Pr(w\cond y=s,d)Pr(y=s),$$
unlike MOU. The LBG model therefore addresses the two main drawbacks of MOU: first being that the single-topic assumption does not apply to the entire document but only to selected keywords, and secondly that the word generation process depends also on the document thereby allowing multiple meanings to words. Both of those "upgrades" make the underlying model assumptions more realistic than MOU. Consequently, the LBG model can be considered as a natural extension of the MOU model where some of the limiting (and unrealistic) assumptions of MOU are relaxed.

The  LDA model addresses the single-topic assumption of MOU by allowing multiple topics per document in the following manner. To generate the $N$ words of a document $d$, (i) a $k$-faced topic-die is generated by sampling from a Dirichlet distribution with parameters $\alpha_1,...,\alpha_k$, then (ii) repeat $N$ times: (a) sample a topic $s$ by tossing the topic-die, and (b) sample a word by tossing the word-die $\bfbeta_s$.

The parameters $\bfalpha,\bfbeta$ of the LDA model are learned through a Variational EM algorithm. Unlike MOU and LBG, in the LDA model the topic is selected {\it per word\/} rather than once per document. This approach definitely solves the single-topic limitation of MOU and also the polysemy issue since the posterior $Pr(y=s\cond w,d)$ depends on $d$:
$$Pr(y=s\cond w,d)\propto Pr(w\cond y=s)Pr(y=s\cond d).$$
However, there is a price to pay for the powerful generality of the LDA model. First, the posteriors $P(y=s\cond d)$ are computationally intractable and instead are replaced by a mean-field "surrogate" approximation or by sampling methods.
Secondly, by design, LDA requires a relatively large number of topics $k$ (around $\sim 50$) which is fine in the world of text but is limiting to other data analysis domains where the number of "topics" are known to be small (like clustering applications).

In practice, LDA is often used for dimensionality reduction (using the variational parameters $\gamma\in R^k$ per document) as a filter for SVM classification and for supervised classification by performing a separate LDA modeling per class. Despite the reservations above, there are situations where the powerful generality  of the LDA model pays off --- in the domain of text this happens when two topics are very similar. In such cases, the modeling capacity of MOU and LBG is too limited and cannot separate the two classes (see sec.~\ref{sec:exp} for details).

The pLSI model represents the training data as a mixture of multinomials and, like LDA, also allows for multiple topics per document. The pLSI model (unlike MOU, LDA and LBG) is not generative, i.e., there is no natural way to use the model to assign probability to a novel document. Related to that, the number of parameters of the model grows linearly with the training set thus risking an over-fitting phenomenon to occur. The pLSI model, therefore, is not a natural candidate for classification tasks because a novel data instance cannot be classified without essentially retraining the entire dataset. We refer the reader to \cite{BleiNJ03} for a detailed comparison between LDA and pLSI. We have included pLSI in our experiments (sec.~\ref{sec:exp}) as one can often obtain good performance if retraining is allowed during classification of a novel document.

\section{Experiments}
\label{sec:exp}

\begin{figure*}
\centering
    \subfigure[Semantically-unrelated topics]{\includegraphics[width=.85\columnwidth]{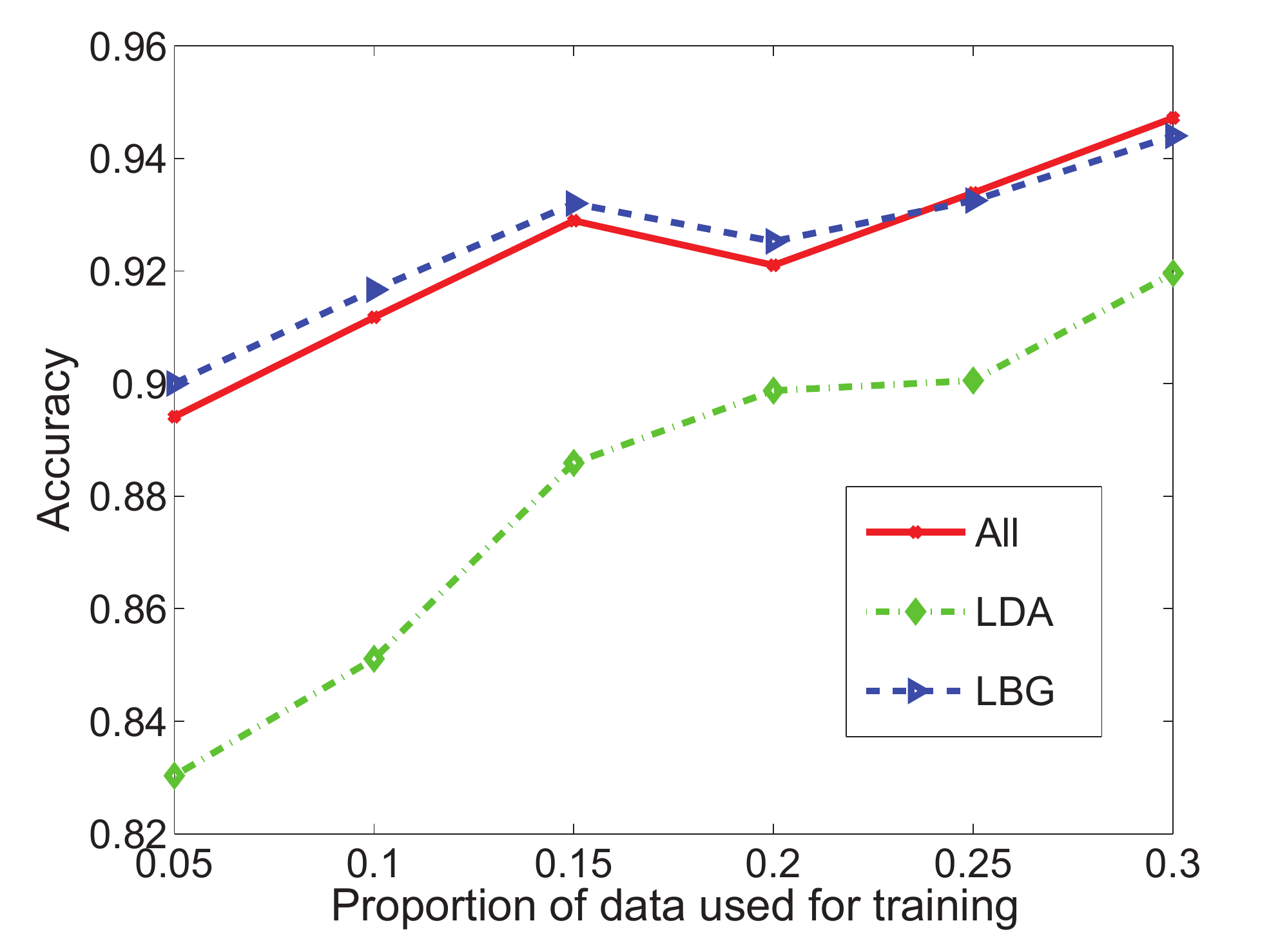}}
    \hspace{.3in}
    \subfigure[Semantically-close topics]{\includegraphics[width=.85\columnwidth]{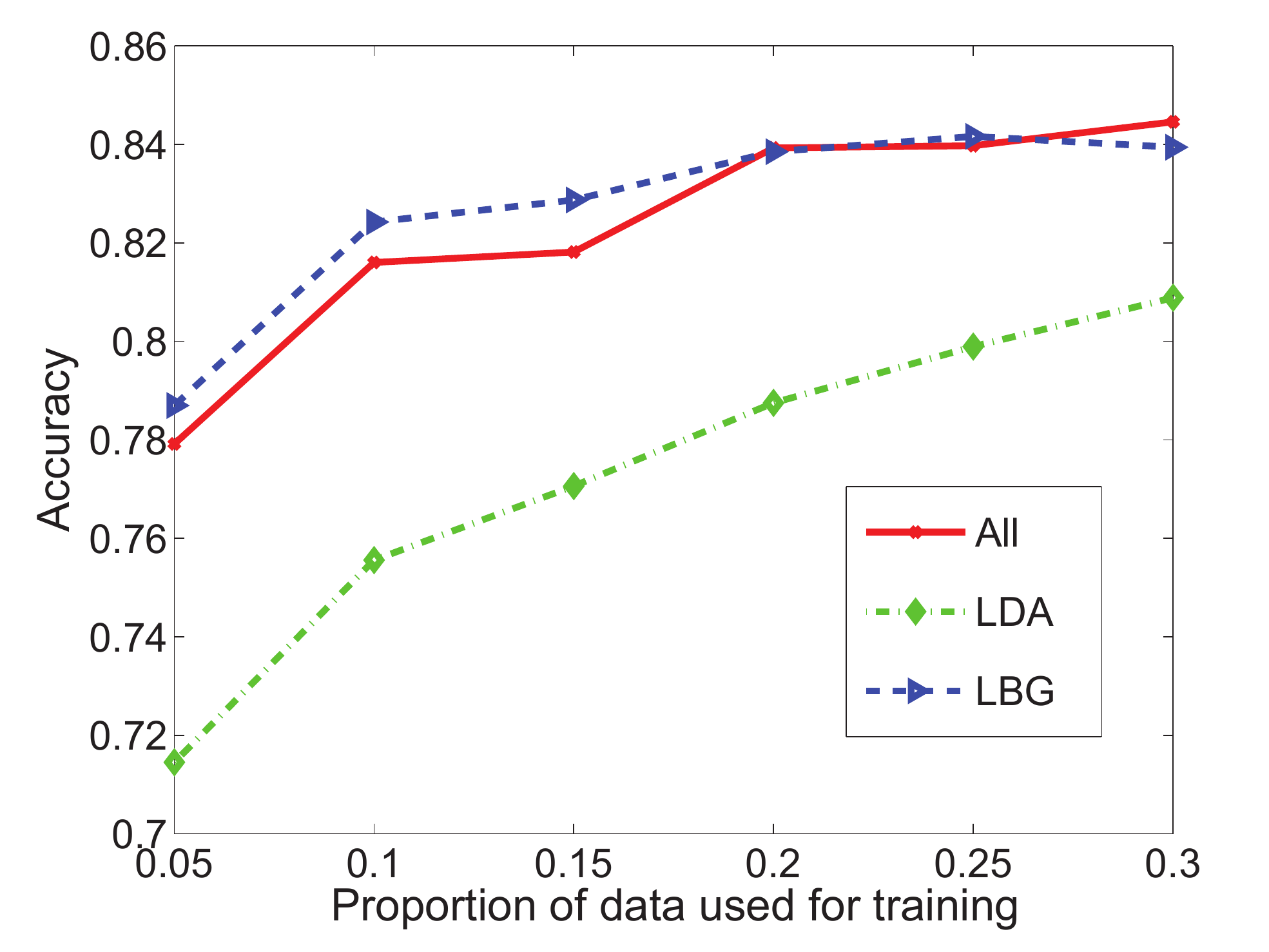}}
	\label{fig:SVM_20NG}
\caption{\small 20NewsGroup classification result on a binary classification problem, using SVM on the reduced set of features. Graph (a) is misc.forsale vs. rec.sport.baseball. Graph (b) is comp.graphics vs. comp.os.ms\_windows.misc. }
\end{figure*}

We conducted experiments with MAP classification, feature selection as a filter for SVM and K-means, supervised classification fitting a model per class and collaborative filtering. Those experiments were conducted on a number of datasets including  20NewsGroup\footnote{The 20NewsGroup data set, taken from the Usenet Newsgroup Collection, consists of some $20,000$ newsgroup postings, each one categorized to a different topic where each topic contains $1000$ documents.}, 100KMovieLens, and Spambase from UCI ML repository.


\noindent {\bf MAP Unsupervised Classification:\ } we begin with an unsupervised classification experiment using the MAP output of our model (the posteriors $\mu_s=Pr(y=s\cond d)$). We randomly split the data set into training and test subsets, generated by mixing records from all the topics in the data set. Having stripped all the record headers, a code-book is created, comprising of all words which are not stop-words in the data set.  We trained a MAP classifier with our model and evaluated the classification output by comparing the cluster label of each record with its true label, as per the 20NewsGroup data set.

In order to measure the clustering performance, we use the zero-one loss function, as follows. Given the $i$'th posting, let $s_i$ and $\kappa_i$ be the obtained cluster label and the true label, respectively. The accuracy (AC), is defined by
$AC=(1/m)\sum_{i=1}^m \delta(\kappa_i,map(s_i))$,
where $\delta(x,y)$ is an indicator function that equals one
if $x=y$ and zero otherwise; and $map(s_i)$ is the permutation mapping function that maps each cluster label $s_i$ to its equivalent label from the data set. The optimal mapping is obtained by  the Kuhn-Munkres algorithm~\cite{Lovasz_MatchingTheory}.
We compared our results with those of MOU, pLSI and LDA. For the latter, a clustering decision was made by examining the $\phi_i$ variational parameters that are introduced for each record. We repeated the experiments several times and the average results are reported in Table ~\ref{tab:20NewsGroup_MAP_classification_All_classes}. Note that pLSI retrains the entire data for each new test record thus skewing the comparison --- yet it is interesting to note that LBG matched the performance nevertheless. Note the large performance gap between LBG and MOU underscoring the significant upgrade to the MOU model. We conjecture that the relatively low accuracy obtained by the LDA model is related to the mean-field approximation and as mentioned above, LDA is hardly ever used for MAP applications for presumably the same reasons.

\begin{table}
    \caption {MAP classification performance comparison for the 20NewsGroup data set.}
    \label {tab:20NewsGroup_MAP_classification_All_classes}
    \vskip 0.15in
	\centering
		\begin{tabular} {|r|r|r|r|}
			\hline
			 LBG & MOU & pLSI & LDA \\
			 \hline
			28\% & 15\% & 27\% & 12\%\\
			 \hline
		\end{tabular}
        \vskip -0.1in		
\end{table}

\noindent{\bf Feature Selection:\ } we compared the performance of our feature selection procedure, as described in sec.~\ref{sec:app}, with the dimensionally reduction offered by the variational parameters $\gamma$ of the LDA model. In our first experiment, we selected a pair of classes  from the 20NewsGroup dataset and performed an SVM classification where the representation of data-instances were the selected coordinates given by LBG or the reduced dimension vector $\gamma$ provided by the LDA model. For control purposes we also applied SVM on the raw representation (without the filter). Fig.~\ref{fig:SVM_20NG} shows the classification accuracy results for two pairs of classes --- a semantically close pair and a pair of unrelated classes. Several experiments were conducted where the proportion of the training data was varied --- from $5\%$ to $30\%$. One can see that the LBG filter produced accuracies comparable to raw data use (slightly better for small training sets) with consistently better performance than the LDA filter\footnote{LDA  at http://chasen.org/$\sim$ daiti-m/dist/lda/}. Note that all approaches suffered when applied to a semantically-related pair of classes.

In the second experiment, we performed an unsupervised classification using K-means clustering on the filtered representations and without the filter (the raw data). Results for both semantically-close and semantically-unrelated pairs of classes are shown in Tables~\ref{tab:sameTopics_Kmeans} and~\ref{tab:differTopics_Kmeans}. One can see that LDA can produce a superior accuracy when the two classes are semantically-close (comp.os.ms\_windows.misc versus comp.graphics). LBG on the other hand consistently outperformed LDA for semantically-unrelated clusters.

\begin{table*}[t]
    \caption {K-means classification for semantically-close classes.}
    \label{tab:sameTopics_Kmeans}
    \vskip 0.15in
	\centering
		\begin{tabular} {|c|c|c|c|c|}
			\hline
			 & comp.os.ms\_windows.misc & talk.politics.mideast & rec.sport.baseball & talk.religion.misc\\
			 & comp.graphics    & talk.politics.misc    & rec.sport.hockey   & talk.religion.cristianity\\
            \hline
			LBG           & 63.25\%  & 72.75\%   & 52.875\% & 57.625\%  \\
			LDA          & 77.5\%   & 57.75\%   & 53\%     & 58.375\%  \\
			All          & 50.25\%  &  58.625\% & 50.375\% &  50.5\%     \\
			\hline
		\end{tabular}
        \vskip -0.1in
\end{table*}

\noindent{\bf Collaborative Filtering:\ }
We used the 100KMovieLens Collaborative-Filtering data, which consists of approximately $100,000$ ratings for $1,682$ movies by 943 viewers. 
As discussed in sec.~\ref{sec:app}, we train our model using a fully-observed set of viewers. Then, for every test viewer, we suppress a single, randomly-chosen movie rating. Our task is to predict the rating, given all the other movies for which that viewer has voted (known as the "Forced Prediction" protocol).
Adopting Hofmann~\yrcite{CF_Hoff03} and Breese~\yrcite{Breese98empiricalanalysis}, we use two evaluation metrics which measure the distance of the estimated vote $\hat{m}$ from the true vote $m$ ---  the mean absolute error MAE, $avg(|\hat{m}-m|)$, and the rooted mean squared error RMSE $avg((\hat{m}-m)^2)$.
We then compared our method to Gassian-pLSA proposed by~\cite{CF_Hoff03} and to the Baseline method that simply outputs the mean vote over the entire training data for each movie. The results are displayed in Table~\ref{tab:CF_Prediction_comparison}. Note that LDA and pLSI do not naturally accommodate the Forced Prediction protocol as they  do not measure word frequencies, thus were omitted from the comparison. One can see that LBG produced a lower MAE error compared to both Gaussian-pLSA and the Baseline method and slightly lower error on the RMSE measure (compared to Baseline).

\begin{table*}[t]
    \caption {K-means classification for semantically-unrelated classes.}
    \label{tab:differTopics_Kmeans}
    \vskip 0.15in
	\centering
		\begin{tabular} {|c|c|c|c|c|}
			\hline
			 & comp.windows.misc\_windows.misc & comp.sys.mac.hardware & alt.atheism & rec.sport.baseball\\
			 & rec.autos                        & alt.atheism          & rec.motorcycles   &  misc.forsale\\
            \hline
			LBG           & 93.75\%  & 97.25\%   & 94.125\% &     93\%\\
			LDA          & 84.125\%   & 90.125\%   & 58.25\%&    89\%  \\
			All          & 95.375\%  &  96.875\% & 88.875\% &    93\%     \\
			\hline
		\end{tabular}
        \vskip -0.1in
\end{table*}

\noindent{\bf Spam Filtering:\ }
The Spambase data set from the UCI Machine Learning Repository  dataset consists of 4601 of emails ("documents"), characterized by 54 attributes ("words") plus a class label ("spam"=positive/"ham"=negative) where $39\%$ of the emails are labeled as spam.
We begin with an unsupervised MAP estimation where  Table~\ref{tab:Spam_unsupervised} displays the performance of LBG against  MOU and pLSI (where with pLSI a retraining is required for each test data). Note the performance gap between LBG and pLSI --- this we conjecture has to do with the plausibility of  the single-topic assumption for spam filtering -- words with high percentage occurrence serve as a natural discriminative indicator (for example, an email with repeated occurrences of the word "buy" is likely to be spam). The performance gap with MOU is attributed to the fact that the single-topic assumption is best applied on keywords rather than on all words of the document.
\begin{table}
    \caption {Unsupervised spam-filter classification performance comparison.}
    \label {tab:Spam_unsupervised}
    \vskip 0.15in
	\centering
		\begin{tabular} {|r|r|r|}
			\hline
			 LBG & MOU & pLSI  \\
			 \hline
			78\% & 60\% & 65\% \\
			 \hline
		\end{tabular}
        \vskip -0.1in
\end{table}
\begin{table}
        \caption {MovieLens Collaborative Filtering prediction results.}
        \label {tab:CF_Prediction_comparison}
        \vskip 0.15in
        \begin{tabular*}{.4\textwidth}%
        {|l|c@{\extracolsep{\fill}}c|}\hline
        Method & \multicolumn{2}{c|}{Absolute Error}\\
        & MAE & RMSE\\\hline
        Baseline         &  0.905  &  1.1445  \\
        Gaussian pLSA    & 1.884   &  2.1142  \\
        LBG  &  0.776  &  1.1183  \\
        \hline
        \end{tabular*}
        \vskip -0.1in
\end{table}

\begin{figure}[h]
\includegraphics[height=0.15\textheight]{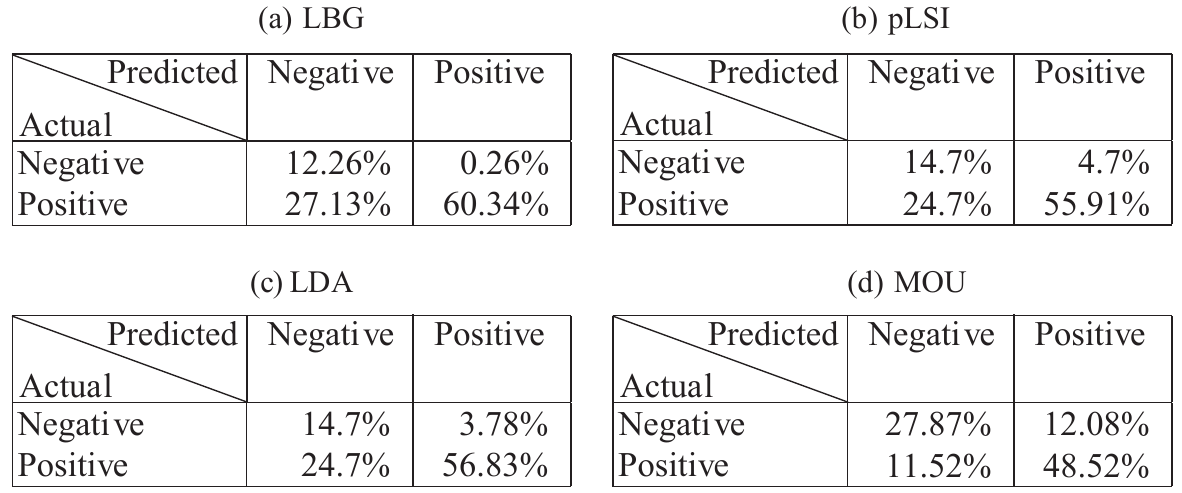}
\caption{Confusion tables for supervised spam-filter.}
\label{fig:spamConfusionTables}
\end{figure}


We then moved to a supervised setting, in which we used the class labels (spam/ham) in the training stage. We modeled each class separately using LBG, LDA, MOU and pLSI while fitting the optimal number of topics per model (see note at the end of sec.~\ref{sec:bg}). Note that MOU, when $k=1$,  reduces to the SpamBayes algorithm.
The confusion table of each method is displayed in Fig.~\ref{fig:spamConfusionTables}.
Note the strikingly low false-positive (ham classified as spam) result for the LBG model, compared to other models. Future work might be directed to the development of an enhanced model, which will compensate for LBG's limited success with false-negatives.

\bibliographystyle{mlapa}

{\small
\bibliography{LBG}

%

\end{document}